# LatentGaze: Cross-Domain Gaze Estimation through Gaze-Aware Analytic Latent Code Manipulation

Isack Lee[†], Jun-Seok Yun[†], Hee Hyeon Kim, Youngju Na, Seok Bong Yoo[*]

Artificial Intelligence Convergence, Chonnam National University, Gwangju, Korea
{sackda24, 218062, sbyoo}@jnu.ac.kr

**Abstract.** Although recent gaze estimation methods lay great emphasis on attentively extracting gaze-relevant features from facial or eyeimages, how to define features that include gaze-relevant components has been ambiguous. This obscurity makes the model learn not only gaze-relevant features but also irrelevant ones. In particular, it is fatal for the cross-dataset performance. To overcome this challenging issue, we propose a gaze-aware analytic manipulation method, based on a data-driven approach with generative adversarial network inversion's disentanglement characteristics, to selectively utilize gaze-relevant features in a latent code. Furthermore, by utilizing GAN-based encoder-generator process, we shift the input image from the target domain to the source domain image, which a gaze estimator is sufficiently aware. In addition, we propose gaze distortion loss in the encoder that prevents the distortion of gaze information. The experimental results demonstrate that our method achieves state-of-the-art gaze estimation accuracy in a cross-domain gaze estimation tasks. This code is available at https://github.com/leeisack/LatentGaze/.

## 1 Introduction

Human gaze information is essential in modern applications, such as human computer interaction [1, 2], autonomous driving [3], and robot interaction [4]. With the development of deep-learning techniques, leveraging convolution neural networks (CNNs), appearance-based gaze estimation has led to significant improvements in gaze estimation. Recently, various unconstrained datasets with wide gaze range, head pose range are proposed. Although these allow a gaze estimator to learn a broader range of gaze and head pose, the improvement is limited to the fixed environment. Because real-world datasets contain unseen conditions, such as the various personal appearances, illuminations, and background environments, the performance of the gaze estimator be degraded significantly. It suggests that gaze-irrelevant factors cause unexpected mapping relations, so called overfitting. It makes a model barely handle various unexpected factors in a changing environment. Therefore, to handle this issue, the domain adaptation method, which uses a small number of target samples in training, and the domain generalization method, which only utilizes source domain data, have received much attention recently. However, in the real world, the former approach does not always hold in practice because the target data for adaptation are usually unavailable. Consequently, the latter approach is highly effective because of its practicability. Especially in gaze estimation, these problems are more challenging beca-use the dimensions of gaze-irrelevant features are much larger than those of gaze-relevant features, which are essential to gaze estimation. This problem must be considered because it affects the learning of the network in an unexpected way. Although a few studies have suggested handling this issue in gaze estimation tasks, it remains a challenge.

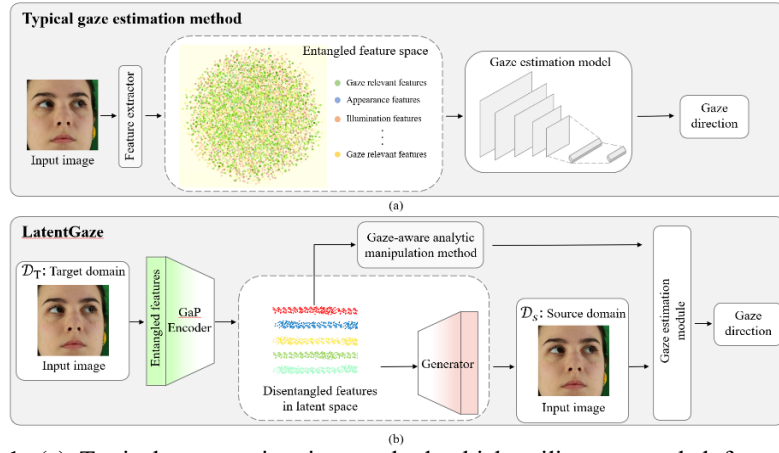

**Fig. 1.** (a) Typical gaze estimation method which utilizes entangled features. (b) LatentGaze which utilizes latent code with domain shifted image.

In this paper, to solve this issue, we propose a gaze-aware latent code manipulation module based on the data-driven approach. In this process, we focus on the following two characteristics of Generative Adversarial Networks (GAN) inversion for downstream tasks: reconstruction and editability. The former characteristic assures that the attributes of the input images can be mapped into a latent code without losing principal information. The latter enables manipulation or editing concerning a specific attribute, allowing the latent code to be selectively used by excluding or remaining the information in downstream tasks. Based on these, our proposed method utilizes statistical analysis in the latent domain to find elements that have a high correlation with the gaze information. To this end, we split the dataset into binary-labeled groups, which are a left-staring group and a right-staring group. We then perform two-sample $t$-tests to see whether the difference between the distributions of two groups' latent codes is statistically significant. Therefore, we could find which elements are related to gaze. In addition, the specific information of the latent code is attentively used to train a gaze estimator. As shown in Fig. 1, typical gaze estimation methods are vulnerable to domain shifts because they use an image with entangled attributes. However, GAN inversion separates intertwined characteristics within an image into semantic units (e.g., illumination, background, and head pose). We exploit this advantage by using the GAN inversion encoder as a backbone for the two modules in our framework. We propose a target-to-source domain shift module with a GAN-based encoder-decoder. This module maps the unseen target image into the source distribution, where the gaze estimator is sufficiently aware, by replacing the attributes of the image with the attributes of the source domain. To this end, we first use the GAN inversion method, encoder, to extract the input image's attributes in a semantically disentangled manner. Nevertheless, while coarse information is generally maintained after inversion, finer information such as g-aze angle and skin tone are subject to distortion. Thus, we introduce our proposed gaze distortion loss in the training process of the encoder to minimize the distortion of gaze between the input and the generated images. Our two modules complementarily improve performance during gaze estimation because they encompass both the image and the latent domains. The contributions of this study can be summarized as follows.

• We propose an intuitive and understandable gaze-aware latent code manipulation module to select semantically useful information for gaze estimation based on a statistical data-driven approach.
• We propose a target-to-source domain shift module that maps the target image into the source domain image space with a GAN-based encoder-decoder. It significantly improves the cross-dataset gaze estimation performance.
• We demonstrate the correlation between manipulated latent features and gaze estimation performance through visualizations and qualitative experiments. As a result, our framework outperforms the state-of-the-art (SOTA) methods in both fixed and cross-dataset evaluations.

## 2 Related Work

### 2.1 Latent Space Embedding and GAN Inversion

Recent studies have shown that GANs effectively encode various semantic information in latent space as a result of image generation. Diverse manipulation approaches have been proposed to extract and control image attributes. At an early stage, Mirza et al. [5] trained to create a conditional image which enables the control of a specific attribute of an image. Subsequently, Abdal et al. [6] analyze three semantic editing operations that can be applied on vectors in the latent space. Shen et al. [7] adopted a data-driven approach and used principal component analysis to learn the most important directions. Park et al. [8] presented a simple yet effective approach to conditional continuous normalizing flows in the GAN latent space conditioned by attribute features. In the present study, we propose an intuitive and understandable manipulation method to find the direction that correlates with gaze estimation performance. In addition, we prove the effectiveness and necessity of latent code manipulation, particularly by showing cross-domain evaluations. However, such manipulations in the latent space are only applicable to images generated from pre-trained GANs rather than to any given real image. GAN inversion maps a real image into a latent space using a pre-trained generator. Inversion not only considers reconstruction performance but must also be semantically meaningful to perform editing. To this end, Zhu et al. [9] introduced a domain-guided encoder and domain regularized optimization to enable semantically significant inversion. Tov et al. [10] investigated the characteristics of high-quality inversion in terms of distortion, perception, and editability and showed the innate tradeoffs among them. Typically, encoders learn to reduce distortion, which represents the similarity between the input and target images, both in the RGB and feature domains. Zhe et al. [11] proposed an architecture that learns to disentangle and encode these irrelevant variations in a self-learned manner. Zheng et al. [12] proposed a method to relieve distortion of the eye region by leveraging GAN training to synthesize an eye i-mage adjusted on a target gaze direction. However, in this study, the distortion is redefined from the gaze estimation perspective. By utilizing GAN inversion as a backbone for two proposed modules, we effectively improve generalization ability of our framework. First, we take the editability of GAN inversion to extract gaze-relevant features and utilize them in gaze-aware analytic selection manipulation module. Second, we improve the estimation performance for unseen datasets by shifting target domain images into source domain images utilizing GAN inversion and generator. It reduces the bias between the distributions of the two independent image spaces. These help the gaze estimator improve cross-domain performance without touching the target samples.

**2.2 Domain Adaptation and Generalization**

Most machine learning methods commonly encounter the domain shift problem in practice owing to the distribution shift between the source and target domain datasets. Consequently, a model suffers significantly from performance degradation in the target domain. Several research fields have been explored extensively to address this problem. Domain adaptation utilizes a few target samples during the training [13]. Recently, Bao et al. [14] performed the domain adaptation under the restriction of rotation consistency and proposed rotation-enhanced unsupervised domain adaptation (RUDA) for cross-domain gaze estimation. Wang et al. [15] proposed a gaze adaptation method, namely contrastive regression gaze adaptation (CRGA), for generalizing gaze estimation on the target domain in an unsupervised approach. Although such methods show performance improvement, they do not always hold in practice as target data are difficult to obtain or are unknown. In contrast, domain generalization aims to generalize a model to perform well in any domain, using only the source domain dataset. Unsupervised domain generalization is primarily based on two methodologies: self-supervised learning and adversarial learning. An advantage of self-supervised learning is that the model learns generic features while solving a pretext task. This makes the model less prone to overfitting to the source domain [16]. Adversarial learning allows a generative model to learn the disentangled representations. Here, we introduce GAN inversion to utilize its out-of-distribution generalization ability.

**2.3 Gaze Estimation**

Several appearance-based gaze estimation methods have been introduced [17-19]. Recent studies have shown significant improvements in gaze estimation performance using various public datasets [20]. However, cross-person and cross-dataset problems remain in gaze estimation tasks. This stems from variance across a large number of subjects with different appearances. To improve the cross-person performance, Park et al. [8] utilized an auto-encoder to handle person-specific gaze and a few calibration samples. Yu et al. [21] improved the person-specific gaze model adaptation by generating additional training samples through the synthesis of gaze-redirected eye images. Liu et al. [22] proposed an outlier-guided collaborative learning for generalizing gaze in the target domain. Kellnhofer et al. [23] utilized a discriminator to adapt the model to the target domain. Cheng et al. [24] introduced domain generalization method through gaze feature purification. This method commonly assumes that gaze-irrelevant features make the model overfit in the source domain and perform poorly in the target domain. Although these methods have improved cross-person performance, they cannot always be applicable as the target data are difficult to obtain or even unavailable in the real world. In consideration of this, generalizing the gaze estimation across datasets is required. However, since facial features are tightly intertwined with illumination, face color, facial expression, it is difficult to separate gaze-relevant features from them. To address this, we improve the generalization ability by analyzing latent codes and selectively utilize the attributes that are favorable to the gaze estimator.

## 3 Method

### 3.1 Preliminaries

Gaze estimation models which use full-faces learn a biased mapping on the image-specific information such as facial expressions, personal appearance, and illumination rather than learn only the gaze-relevant information of the source dataset. While it leads to a strong mapping ability in a fixed domain, it still remains as a challenging problem as it causes serious performance degradation in cross-domain gaze estimation. Therefore, this paper aims to dodge the bias-variance tradeoff dilemma by shifting the domain of image space from target dataset to source dataset. In this section, we introduce some notations, objective definitions and condition.

**Conditions.** While the majority of generalization works have been dedicated to the multi-source setting, our work only assumes a single-source domain setting for two motivations. First, it is easier to expand from a single-source domain generalization to a multi-source domain generalization. Second, our study specifically attempts to solve the out-of-distribution problem when the training data are roughly homogeneous.

**Definition of Latent Code Manipulation.** The first application of GAN inversion is to selectively utilize gaze-related attributes in the input image. We invert a target domain image into the latent space because it allows us to manipulate the inverted image in the latent space by discovering the desired code with gaze-related attributes. This technique is usually referred to as latent code manipulation. In this paper, we will interchangeably use the terms "latent code" and "latent vector". Our goal is to ensure that the manipulated latent vector and gaze vector are highly correlated. Therefore, we need to find the manipulation operator $H$ that determines which elements are correlated to gaze information in the latent vector. This objective can be formulated as:

$$H^* = \max_{H} \mathcal{F}(H(X)), \quad (1)$$

$$\mathcal{F} = \left[\frac{\partial f}{\partial x_1} \frac{\partial f}{\partial x_2} \cdots \frac{\partial f}{\partial x_d}\right]^T, \quad (2)$$

where $f$ denotes an optimal gaze estimator, $X$ denotes latent vector. $x$ denotes an element of $X$, and $d$ denotes the dimension of latent space.

**Definition of Domain Shifting Process.** Given labeled source domain $D_S$, the goal is to learn a shifting model $f$ that maps an unseen image from target domain $D_T$ to $D_S$ by only accessing source data from $D_S$. We take GAN inversion method into consideration to utilize its out-of-distribution generalization ability. It can support inverting the real-world images, that are not generated by the same process of the source domain training data. Our objective can be formulated as follows:

$$S^* = \arg\min_{S}\|F(S(D_T) - F(D_S))\|, \qquad (3)$$

Where $F$ denotes the general feature extractor, and $S^*$ denotes the domain shifter. Note that domain the generalization process should not include the target sample during the training. We utilize a StyleGAN-based encoder-generator to shift the domain.

### 3.2 The LatentGaze Framework Overview

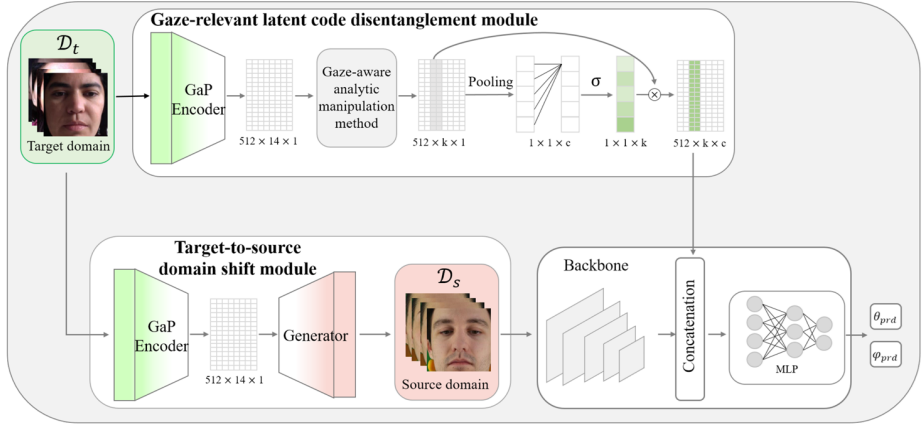

**Fig. 2.** Overview of the LatentGaze framework. Our framework is mainly composed of two modules. The target-to-source domain shift module maps the unseen target image into the source distribution where the gaze estimator is sufficiently aware. The gaze-aware analytic selection manipulation module finds the gaze-relevant feature through statistical analysis in the latent domain. Thereafter, the gaze estimate model utilizes each feature to estimate the gaze vector ($\boldsymbol{\theta}_{prd}$, $\boldsymbol{\varphi}_{prd}$).

Fig. 2 shows the overall architecture of the proposed framework. It consists of two complementary modules. The first module, i.e., the target-to-source domain shift module, maps a target domain image to a source domain image. The encoder extracts an image into a latent code. To reduce the possible gaze distortion in the encoder, we propose gaze distortion loss to preserve the gaze information. The generator generates an attributes-persevered image in the source domain, based on the extracted latent code. The second module, i.e., the gaze-aware latent code manipulation module, improves the generalization performance by manipulating features that are highly related to gaze in the latent code through statistical hypothesis test. Subsequently, the latent vector and the image are concatenated. And through two fc-layers, the gaze direction is predicted as $\boldsymbol{\theta}_{prd}$ and $\boldsymbol{\varphi}_{prd}$. These two modules effectively utilize the property that latent code passed through inversion is semantically disentangled. These two modules achieve complementary benefits in our framework by jointly reinforcing cross-domain generalization. Qualitative and quantitative experiments and visualizations show that our framework achieves robust performance, on par with the state-of-the-art approaches.

### 3.3 Statistical Latent Editing

Recently, extensive research has been conducted on the discovery of semantically meaningful latent manipulations. Radford et al. [25] investigated the vector arithmetic property in the latent space of a pre-trained generator and demonstrated that averaging the latent code of samples for visual concepts showed consistent and stable generations that semantically obeyed arithmetic. Therefore, we assume that it is possible to statistically find gaze-related features that correspond to a specific attribute. In this study, we propose an explainable and intuitive gaze-aware analytic manipulation method to select semantically useful information for gaze estimation based on a statistical data-driven approach. We utilize ETH-XGaze [26] and MPIIFaceGaze [27] which have a large number of subjects, wide gaze and head pose ranges for the analysis. This method consists of four steps. First, we divide the data set into two groups $G_L$ and $G_R$. $G_L$ denotes the group having gaze vectors of 30 degrees to 90 degrees along the yaw axis direction. $G_R$ denotes the group having gaze vectors of -30 degrees to -90 degrees along the yaw axis direction. Second, we calculate the mean of the entire latent codes for each group. Since gaze-irrelevant attributes such as illumination, person-appearance, background environments are averaged out, the divided groups have the same value of latent codes except for gaze-relevant elements. We can presume that the elements with a large difference between the group's values are significantly related to gaze features. From then on, since adjacent tensors represent similar features, we consider 16 tensors as a single chunk and define it as a unit of statistical operations as follows.

$$C^i = \frac{1}{16}\sum_{j=1}^{16} C^{ij}, \ i = 1, 2, \dots, k, \tag{4}$$

where $C^i$ denotes $i$-th chunk, $k$ denotes the number of chunks in a latent code, and $j$ represents the number of elements in a chunk. $C_L^i$ and $C_R^i$ each denotes the chunk of the group $G_L$ and $G_R$. Third, we sort the chunks in a descending order to select the most gaze-related chunks. Fourth, we perform paired *t*-tests for two expected values of each chunk from two groups.

**Paired *t*-test for Two Expected Values.** We perform the *t*-test and the size of the samples in the two groups is sufficiently large, so the distribution of the difference between the sample means approximates to Gaussian distribution by the central limit theorem, and each sample variance approximates to the population variance. The hypotheses and test statistic are formulated as:

$$H_0 : \mu_L^i = \mu_R^i, \ H_1 : \mu_L^i \neq \mu_R^i, i = 1, 2, \dots d, \tag{5}$$

$$T = \left(\overline{X}_L^i - \overline{X}_R^i\right) / \sqrt{\frac{\sigma(X_L^i)}{n_L} + \frac{\sigma(X_R^i)}{n_R}}, \tag{6}$$

where $H_0$ denotes null hypothesis, $H_1$ denotes alternative hypothesis, $\mu_L^i$ and $\mu_R^i$ denote the population average of both groups for the $i$-th chunk, $d$ denotes the dimension of the latent space, $T$ denotes test statistic, $X_L^i$ and $X_R^i$ each denotes sample statistic of $C_L^i$ and $C_R^i$, $n_L$ and $n_R$ each denotes to the number of $G_L$ and $G_R$, and $\sigma$ denotes standard deviation. The above procedure can be found in Fig. 3. We found that a number of chunks belonging to the critical region are in the channels 4 and 5. However, as the indices of the gaze-related chunks subject to slightly differ depending on the dataset, we utilize a channel attention layer (CA-layer) [28] to improve generalization performance in cross-domain.

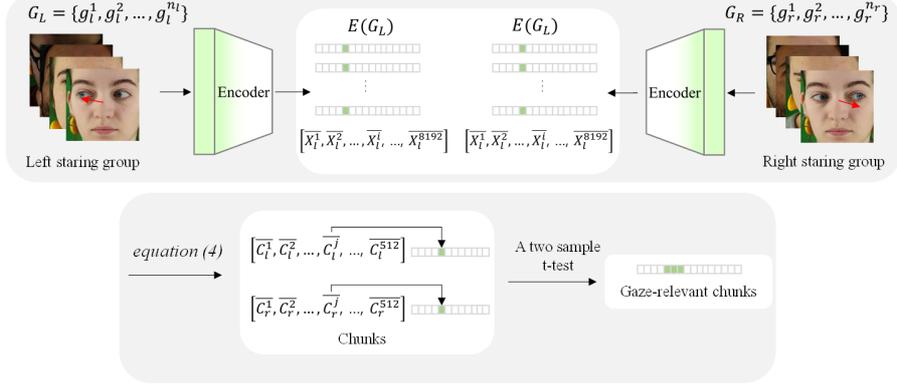

**Fig. 3.** Overview of the statistical latent editing. The person-paired gaze datasets are divided into the left staring group ($G_L$) and right staring group ($G_R$). Each image is embedded into latent code by the GaP-encoder. After that, the latent code values of each group averaged and subtracted the mean values of $G_L$ and $G_R$. The larger values of the result have a high correlation with the gaze information. Due to this statistical hypothesis test, we classify the gaze-relevant and gaze-irrelevant.

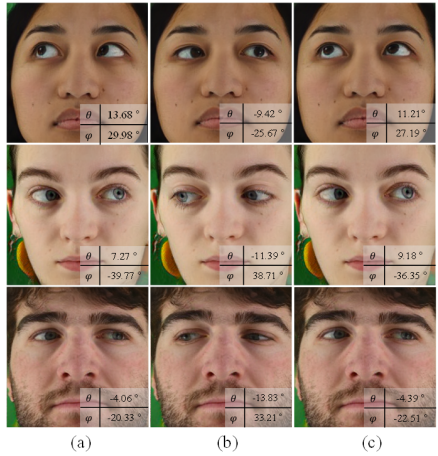

**Fig. 4.** Visualizations of latent code manipulation. (a) shows images staring one side. (b) shows images staring the other side in the same subjects. (c) shows the generated images with the latent codes that combine the gaze-irrelevant chunks of (b) and the gaze-relevant chunks of (a).

We verify the efficacy of our proposed manipulation method through visualizations and experiments. Furthermore, we show the effectiveness of our gaze-aware analytic manipulation method. As shown in Fig. 4, it shows the chunks extracted from our manipulation method are related to gaze information The Fig. 4(c) shows the generated images from the latent code that are replaced only gaze-relevant chunks from the other group. They are similar to the appearance of the Fig. 4(b) while preserving the gaze direction of the Fig. 4(a). Since it is not completely disentangled, it does not completely maintain the appearance, illumination of Fig. 4(b), but it does not affect the performance of gaze estimation.

### 3.4 Domain Shift Module

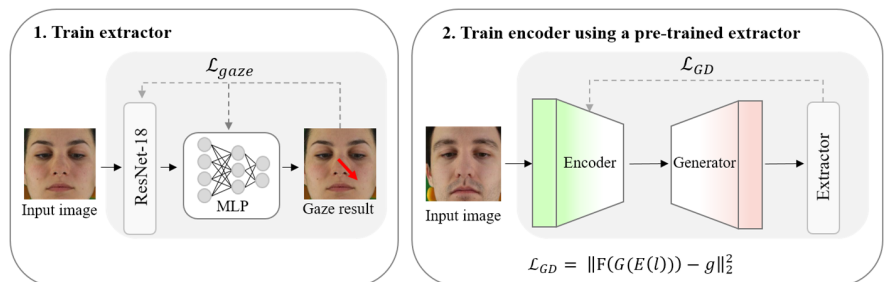

**Fig. 5.** Training process of GaP-encoder that preserves gaze-relevant features during domain shifting. The training process consists of two phases. First, the gaze-relevant extractor is trained to estimate the gaze vector from the facial image. Second, the GaP-encoder is trained by $L_{GD}$ that is difference between the estimated result from generated image $F(G(E(I)))$ and ground truth.

The goal of this module is to properly map a target image to the source latent space by an out-of-distribution generalization of GAN inversion. Latent code should be mapped to a semantically aligned code based on the knowledge that emerged in the latent space. However, due to the large bias of distributions between the source and target datasets, the mapping ability of the encoder network hardly covers the target domain. Furthermore, general extraction of features from the entire image makes it extremely difficult in that it requires extracting features from infinitely large unseen real-world image space. Therefore, to minimize the space of the space that unseen images could possibly have, we capture only face with RetinaFace [29] and crop images. This could drastically reduce the unexpected input from different environments. Since the values of a latent code is set along one specific direction to the corresponding attribute, the latent code is meaningful beyond simply compressing the facial information. The second goal of the generator is to perform image reconstruction at the pixel level. The generator is pre-trained with the source dataset. This makes it possible to utilize dataset-specific mapping ability as the generated images in the space where the gaze estimator is fully capable. Especially, the label space of the gaze estimation task remains the same as it can be represented with two angles of azimuth and elevation. In general, since the generatable space of each attrib-ute trained with the source domain is larger, the generator can reconstruct an image domain from the latent domain with attributes aligned in a semantical way. To obtain latent codes, we benchmarked e4e-encoder [10] which embed images into latent code with disentanglement characters. However, e4e-encoder cannot preserve gaze-relevant features, since the gaze-relevant feature is a relatively finer attribute. To tackle this issue, we propose novel gaze distortion loss $L_{GD}$ that preserves gaze features in the encoding process. We call the encoder with this loss the GaP-encoder. As shown in Fig. 5, training process of encoder consists of two phases. First, we train extractor $F$ that is combined with ResNet-18 and MLP to estimate gaze vector from face images. Second, we utilize $F$ to calculate $L_{GD}$ from the generated image $G(E(I))$. $E$ is an encoder, $G$ is a generator, $I$ is input image, respectively. Therefore, $L_{GD}$ is denoted as:

$$L_{GD} = \|F(G(E(I)) - g\|_2^2, \qquad (7)$$

where $g$ denotes 3D ground-truth gaze angle. The encoder is trained with our proposed $L_{GD}$ on top of three losses used in [10] that consists of $L_2$, $L_{LPIPS}$ [30], $L_{sim}$ [10] and total loss is formulated as follows:

$$L_{total}(I) = \lambda_{l2}L_2(I) + \lambda_{LPIPS}L_{LPIPS}(I) + \lambda_{sim}L_{sim}(I) + \lambda_{GD}L_{GD}(I), \quad (8)$$

where $\lambda_L$, $\lambda_{LPIPS}$, $\lambda_{sim}$, $\lambda_{GD}$ are the weights of each loss. In our method, the extractor $F$ may have gaze angle error. Since our method performs end-to-end training with encoder and extractor $F$ during training process, $F$ can be optimized to generate the image which preserved gaze relevant features.

## 4 Experiments
### 4.1 Datasets and Settings

**Datasets**. Based on the datasets used in recent gaze estimation studies, we used four gaze datasets, ETH-XGaze [26], MPIIFaceGaze [27], and EyeDiap [31]. The ETH-XGaze dataset provides 80 subjects (i.e., 756,540 images) that consist of various illumination conditions, gaze and head pose ranges. The MPIIFaceGaze provides 45000 facial images from 15 subjects. The EyeDiap dataset provides 16,674 facial images from 14 participants. The smaller the number of subjects and various gaze direction datasets, the better to extract gaze-relevant features through statistical methods. For this reason, datasets such as ETH-XGaze and MPIIFaceGaze are useful for extracting gaze-relevant features.

**Cross-Domain Evaluation Environment.** We used ETH-XGaze that has wider gaze and head pose ranges than the others. To evaluate cross-dataset performance of gaze estimation, MPIIFaceGaze and EyeDiap are used. All facial images are resized to $256 \times 256$ for stable encoder-decoder training. We conducted two cross-dataset evaluation tasks, E(ETH-XGaze) → M (MPIIGaze), E → D (EyeDiap).

**Dataset Preparation.** We utilized RetinaFace [30] to detect the face region of an image and consistently cropped according to the region for all datasets to induce the generator to generate equal size of faces.

**Evaluation Metric.** To numerically represent the performance of the gaze estimation models, we computed the angular error between the predicted gaze vector $\hat{g}$ and 3D ground-truth gaze angle $g$, and the angular error can be computed as:

$$L_{angular} = \frac{\hat{g} \cdot g}{\|\hat{g}\|\|g\|}, \tag{9}$$

where $\hat{g}$ and $g$ each denotes predicted 3D gaze angle and the ground truth angle, and $\|g\|$ denotes absolute value of $g$, and $\cdot$ is the dot product.

**4.2 Comparison with Gaze Estimation Models on Single Domain**

**Table 1.** Performance comparison with gaze estimation models for ETH-XGaze and MPIIFaceGaze dataset.

| Method | ETH-XGaze angular error (°) | MPIIFaceGaze angular error (°) |
|---|---|---|
| ResNet-18 | 4.71 | 5.14 |
| Dilated-Net | 4.79 | 4.82 |
| Pure-Gaze | 4.52 | 5.51 |
| RT-Gene | 4.21 | 4.31 |
| LatentGaze | 3.94 | 3.63 |

An experiment was conducted to verify the performance of the gaze estimation model in a single dataset. We achieved favorable performance against other gaze estimation methods on single-domain as well as cross-domain evaluations. We utilized the ETH-XGaze and MPIIFaceGaze dataset which provides a standard gaze estimation protocol on this experiment. ETH-XGaze is divided into 15 validation sets and the rest of the training sets. MPIIFaceGaze is divided into 36,000 training sets, 9,000 validation sets based on each individual. To compare with the SOTA gaze estimation model, we tested RT-Gene, Dilated-Net models using eye images, and ResNet-18, Pure-Gaze using face images. As shown in Table 1, the proposed LatentGaze shows a lower angular error than most others, indicating the superiority of our framework. In addition, as shown in Fig. 6, personal appearances are converted into favorable conditions for gaze estimation.

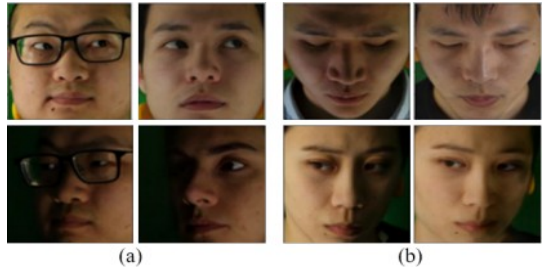

**Fig. 6.** Visual results of LatentGaze in single domain. (a) The glasses are eliminated without blurring and geometric deformation. (b) The illumination is changed favorable to the gaze estimator.

Since it helps the gaze estimation model extract high-quality gaze-relevant features, the model can provide good gaze estimation performance in single domain. In addition to the SOTA performance of our model on the single domain evaluation, we propose a specialized model to cover the cross-domain evaluation.

**4.3 Comparison with Gaze Estimation Models on Cross Domain**

We compare LatentGaze with the SOTA gaze estimation methods on cross-domain evaluation. In general, in a cross-dataset environment, gaze estimation methods show significant performance degradation due to overfitting to the source domain. However, our model solves the overfitting of gaze-irrelevant features by enabling selective latent codes utilization through gaze-aware analytic manipulation. In addition, the target-to-source domain shift module maps the input image to an image space where the gaze estimator is sufficiently aware. As shown in Table 2, our method shows the best or favorable gaze angular error in two cross-domain environments. Here, the performance of E→D is poor compared to E→M. The reason is that the difference in resolution between the source domain and the target domain used for generator training is extreme. However, we can solve this through preprocessing through super-resolution and denoising. In this experiment, no more than 100 target samples for adoption were randomly selected. It demonstrates that our framework is suitable for a real-world application where the distribution of target domain is usually unknown. Moreover, as shown in Fig. 7, it shows that the proposed framework can shift the target domain to the source domain while preserving gaze-relevant features. Since it helps the gaze estimation module maintains gaze estimation performance, our framework provides the robustness of cross-domain evaluation. Even if small illumination change is generated during the domain shift process as shown in Fig. 7, it does not affect the gaze estimation performance as shown in Table 2.

**Table 2.** Performance comparison with SOTA gaze estimation models on cross-dataset evaluation. The best and the second-best results are highlighted in red and blue colors, respectively.

| Method | Target samples | E → M | E → D |
|---|---|---|---|
| Gaze360 [33] | > 100 | 5.97 | 7.84 |
| GazeAdv [34] | ∼ 500 | 6.75 | 8.10 |
| DAGEN [16] | ∼ 500 | 6.16 | 9.73 |
| ADDA [35] | ∼ 1000 | 6.33 | 7.90 |
| UMA [36] | ∼ 100 | 7.52 | 12.37 |
| PnP-GA [22] | < 100 | 5.53 | 6.42 |
| PureGaze [24] | < 100 | 5.68 | 7.26 |
| CSA [15] | < 100 | 5.87 | 5.95 |
| RUDA [14] | < 100 | 5.70 | 6.29 |
| LatentGaze | N/A | 7.98 | 9.81 |
| LatentGaze | < 100 | 5.21 | 7.81 |

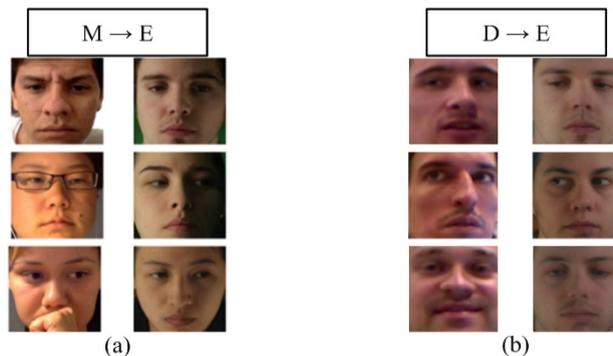

**Fig. 7.** Visualizations of generated images by target-to-source domain shift module. The results show that our module maps the unseen target images into the source distribution by replacing the image's attributes with the source domain's. (a) MPIIFaceGaze dataset is mapped into the domain of the ETH-XGaze dataset. (b) EyeDiap dataset is mapped into the domain of the ETH-XGaze dataset.

**4.4 Ablation Study**

In this section, we conduct experiments about how two modules affect the gaze estimation performance. We also show the necessity of our $L_{GD}$ in the encoder training process by comparing the generated images with and without the loss. Finally, to show the effectiveness of the gaze-aware analytic manipulation method, we first find out the gaze-related index in latent codes through the proposed method and replace one group's gaze-related elements from the other group's ones. And we visualize the generated images from the replaced latent codes to show the gaze of the images are replaced accordingly.

**Effect of Gaze-aware Analytic Manipulation.** In this section, we demonstrate the effectiveness of the gaze-aware analytic manipulation. We used both an image and the manipulated latent code for the MPIIFaceGaze dataset. We used 448 (all), 256 (56%), and 64 (14%) chunks in the latent code, respectively. We removed gaze-irrelevant chunks found by our gaze-aware analytic manipulation method. As presented in Table 3, the angular error using gaze-aware analytic manipulation is lower than that when using many chunks. It indicates that our manipulation method correctly separates gaze-relevant features from tightly intertwined ones. Consequently, the gaze estimation model does not have to consider features that impede the gaze estimation performance.

**Table 3.** Quantitative results for evaluating the effects of gaze-aware analytic manipulation.

| # of Chucks | MPIIFaceGaze angular error(°) |
|---|---|
| None (only ResNet-18) | 5.14 |
| All | 11.34 |
| 256 | 7.21 |
| 64 | 3.63 |

**Effect of GaP-encoder.** To verify the effectiveness of the Gap-encoder, we compare the e4e-encoder benchmarked in this paper with the GaP-encoder. To fair comparison, we train each encoder with 30 epochs, and the angular error is measured by the same gaze estimation module. Also, each image result is generated by the same generator. As presented in Table 4, the GaP-encoder exhibits better gaze estimation performance because our model preserves gaze-relevant features after the domain shifting. As shown in Fig. 8, since our GaP-encoder effectively embed the image into latent code with preserving gaze-relevant features, the generated image shows the same pupil position which is highly related to gaze estimation performance.

**Table 4.** Performance comparison with encoder models. This is the result of using only the latent code from each encoder.

| Method | MPIIFaceGaze Angular error(°) |
|---|---|
| E4e-encoder | 28.51 |
| GaP-encoder | 3.63 |

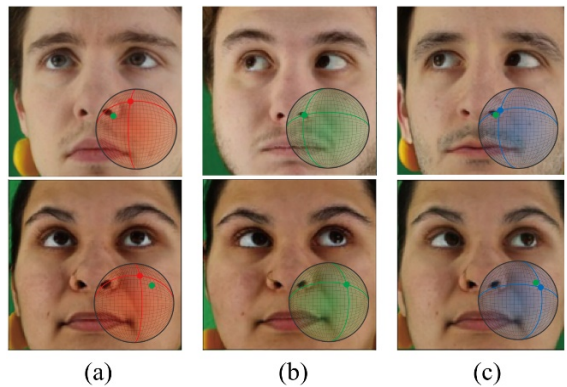

**Fig. 8.** Visual comparison with e4e-encoder and GaP-encoder. All results are generated by the same generator. (a) Images generated by e4e-encoder's latent codes. (b) The original images, (c) Images generated by GaP-encoder's latent codes. Red and blue dots represent the predicted gaze vectors. Green dot represents ground-truth gaze vector.

## 5 Conclusion

In this paper, we presented the first practical application of GAN inversion to solve the cross-dataset problem in gaze estimation with latent code. Our proposed LatentGaze framework consists of a target-to-source domain shift module, which maps the target image into the source domain image space, and a gaze-aware analytic selection manipulation module, which selectively manipulates gaze-relevant features by a statistical data-driven approach. Furthermore, we propose gaze distortion loss that prevents the distortion of gaze information caused by inversion. Our quantitative and qualitative experiments and visualizations show our approach performs favorably against the SOTA methods.